\documentclass{article}

\usepackage[utf8]{inputenc}
\usepackage{amsmath}
\usepackage{graphicx}
\usepackage{hyperref}
\usepackage{geometry}
\usepackage{multicol}
\usepackage{setspace}
\usepackage{booktabs}
\usepackage{authblk}
\usepackage{array} 

\newcolumntype{L}[1]{>{\raggedright\arraybackslash}p{#1}}

\title{An N-of-1 Artificial Intelligence Ecosystem for Precision Medicine}

\author[1]{P. Fard}
\author[1,2]{A. Azhir}
\author[3]{N. Rezaii}
\author[1]{J. Tian}
\author[1]{H. Estiri\thanks{Corresponding author: hestiri@mgh.harvard.edu}}

\affil[1]{Clinical Augmented Intelligence, Massachusetts General Hospital, Boston, MA 02114, USA}
\affil[2]{Department of Medicine, Brigham and Women's Hospital, Boston, MA 02114, USA}
\affil[3]{Department of Neurology, Massachusetts General Hospital, Boston, MA 02114, USA}
\date{}

\begin{document}

\maketitle

\begin{abstract}
Artificial intelligence in medicine is built to serve the average patient. By minimizing error across large datasets, most systems deliver strong aggregate accuracy yet falter at the margins: patients with rare variants, multimorbidity, or underrepresented demographics. This \textit{average patient fallacy} erodes both equity and trust. We propose a different design: a multi-agent ecosystem for N-of-1 decision support. In this environment, agents clustered by organ systems, patient populations, and analytic modalities draw on a shared library of models and evidence synthesis tools. Their results converge in a coordination layer that weighs reliability, uncertainty, and data density before presenting the clinician with a decision-support packet: risk estimates bounded by confidence ranges, outlier flags, and linked evidence. Validation shifts from population averages to individual reliability, measured by error in low-density regions, calibration in the small, and risk--coverage trade-offs. Anticipated challenges include computational demands, automation bias, and regulatory fit, addressed through caching strategies, consensus checks, and adaptive trial frameworks. By moving from monolithic models to orchestrated intelligence, this approach seeks to align medical AI with the first principle of medicine: care that is transparent, equitable, and centered on the individual.
\end{abstract}

\section*{Summary}
\subsection*{Background}
Medicine has always wrestled with the averages. Clinical guidelines are forged from populations, yet the patient before us rarely conforms to that statistical mean. Artificial intelligence has accelerated this tension. Models tuned to minimize average error excel in aggregate performance but stumble on the edge cases, the rare variant, the multi-morbid older adult, and the underrepresented demographic. We have called this the \textit{average patient fallacy}: the quiet but consequential failure of population-trained systems to provide reliable care to the individual.

\subsection*{Methods}
We outline a different architecture: not a single monolithic model but an ecosystem of cooperating agents. Each agent reflects a domain of practice, organs, populations, and modalities, and draws from a shared library of tools: predictors, outlier detectors, uncertainty estimators, and evidence synthesizers. Their outputs converge in a coordination layer that reconciles disagreement, highlights uncertainty, and constructs a packet of decision support for the clinician: risk estimates with confidence bounds, outlier signals, and curated evidence. In this design, validation shifts from aggregate statistics to individual reliability, measured by error in low-density regions, calibration in the small, and the ability to abstain when knowledge is thin.

\subsection*{Findings}
This agentic ecosystem is engineered to surface what single models suppress. By distributing intelligence across specialties, weighting it by reliability, and making uncertainty visible, the system acknowledges what medicine has always known: atypical cases demand extra scrutiny. Failure modes such as outlier neglect, spurious literature synthesis, and overconfidence are countered by explicit safeguards, abstention, consensus checks, and human oversight. The approach anticipates the practical hurdles of computation, regulation, and workflow, and suggests concrete strategies: caching and edge processing to tame latency, adaptive trials to accommodate individualization, and post-market monitoring to ensure safety.

\subsection*{Interpretation}
The lesson is simple but profound: averages do not treat patients, clinicians do. To overcome the average patient fallacy, AI must move from population optimization to individual reliability. A multi-agent ecosystem that quantifies its uncertainty, flags its blind spots, and collaborates with the clinician can make that possible. This is not merely a technical reconfiguration; it is a moral alignment of medical AI with the ethos of medicine itself, care that is transparent, equitable, and centered on the person in front of us.

\subsection*{Summary Panel}
\begin{spacing}{0.9}
\begin{multicols}{2}
\noindent
\textbf{What is already known on this topic}
\begin{itemize}
  \item Artificial intelligence in medicine is largely optimized for population-level accuracy.
  \item Such optimization produces blind spots, especially for patients with atypical presentations, rare conditions, or marginalized backgrounds.
  \item Existing fairness frameworks and multi-agent systems address group disparities or task efficiency but rarely guarantee reliability at the level of the individual.
\end{itemize}

\columnbreak

\textbf{What this framework adds}
\begin{itemize}
  \item We define the \textit{average patient fallacy}: the assumption that models tuned for the mean can serve every patient equally well.
  \item We propose a novel multi-agent ecosystem, in which profession-specialized AI agents draw from a shared model repository and are coordinated to provide N-of-1 decision support.
  \item We outline a roadmap for validation, retrospective analyses, digital twin simulations, and prospective studies that makes individualized AI both testable and tractable.
  \item We argue for a regulatory and ethical framework that measures success not in average accuracy but in fidelity to the single patient at hand.
\end{itemize}
\end{multicols}
\end{spacing}

\section{Introduction}
Medical AI has grown rapidly, from just two FDA-authorized devices in 2015 to over 1,200 by mid-2025. These systems now detect breast cancers, screen for diabetic retinopathy, and recognize subtle patterns invisible to the human eye. Their power comes from training scale: optimizing predictions across large populations.

Yet this is also their weakness. Built for the ``average'' patient, they falter for those at the margins, racial minorities, older adults, or individuals with rare presentations.\cite{ref1,ref2} As coined by Azhir et al.,\cite{ref3} we call this the average patient fallacy: the mistaken belief that models tuned to populations can be trusted equally for every individual.\cite{ref4,ref5} Clinical failures have made this plain, from pulse oximeters misreading oxygenation in patients with darker skin to sepsis models underperforming in older adults.\cite{ref6,ref7,ref8}

Medicine, however, cares for the singular, not the aggregate. The challenge is to build AI that preserves population insights while tailoring predictions to the individual at the point of care.

We propose a new architecture: an N-of-1\cite{ref9,ref10,ref11} ecosystem of specialized AI agents. These agents, clustered by clinical domain or analytic task, draw on a shared repository of predictive models and evidence synthesis tools. A coordination layer integrates their interactions and compiles outputs into tailored decision support, offering clinicians risk estimates, uncertainties, and evidence aligned with the patient before them. This ecosystem is designed not to eliminate averages, but to overcome their tyranny, transforming medical AI from a tool for populations into a companion for individuals.

\subsection{Background}
At the heart of modern medical AI lies a paradox, where AI systems are trained to minimize error across entire populations, yet the very act of optimizing for the majority ensures that some individuals, those whose biology, environment, or presentation lies outside the statistical center, will be poorly served. We call this the average patient fallacy. It is the implicit assumption that a model tuned to maximize accuracy in the aggregate can reliably serve each patient in the clinic. The fallacy emerges whenever a prediction that works well for the many fails the one: when a risk score calibrated on thousands overlooks the atypical physiology of an older adult, or when a dermatology model trained predominantly on lighter skin tones misses a malignant lesion in a patient with darker skin.\cite{ref12,ref13} This is not a new dilemma in medicine. A century ago, drug dosing followed the rule of the ``average man,'' a construct derived from military recruits and textbook physiology. The result was predictable: dosages that overdosed the frail and underdosed the robust. Over time, pharmacology adapted with weight-based dosing, therapeutic drug monitoring, and, more recently, pharmacogenomics, each an attempt to reconcile averages with individuals.\cite{ref14,ref15,ref16} To this date, a constant value of total body oxygen consumption is used for all patients regardless of size, age and sex to calculate Fick-based cardiac output in patients with cardiogenic shock in cardiac care unit. AI now stands at a similar juncture. The models may be novel, but the underlying problem, mistaking the crowd for the person, is not.

The average patient fallacy is a byproduct of how predictive models are trained. A supervised model seeks to minimize the expected squared error:
\[
E[(y - f(x))^2] = \mathrm{Bias}^2 + \mathrm{Variance} + \sigma^2
\]
where $f(x)$ is the model’s prediction for outcome $y$, and $\sigma^2$ represents irreducible noise. When training data are dense and representative, minimizing this error yields reliable estimates. But for patients whose data lie in sparse regions of the feature space, the expected error conditional on that patient, $E[(y - f(x_i))^2 \mid x_i]$, tends to spike. In these low-density regions, the model’s learned associations have low mutual information $I(X;Y)$ with the true outcomes. The model is well-calibrated for the ``center of mass'' of the data distribution, but increasingly unreliable for those on the periphery. Clinical outliers inhabit exactly those regions where information is thinnest and error is highest.

If the average patient fallacy is the blind spot of today’s medical AI, then the solution is not another, larger monolith. It is an ecosystem of specialized agents that reason together and defer when knowledge is thin. The rest of this paper lays out the evidence for why such a system is needed, describes its architecture, and outlines a roadmap for validation and regulation.

\section{Current Multi-Agent Architectures}
Multi-agent systems (MAS) excel in domains where problems decompose cleanly into stable subtasks, agents communicate over reliable channels, and local objectives aggregate into a coherent global goal. Healthcare violates each of these assumptions. A patient's presentation is non-stationary, multimodal, and cross-cutting across organ systems and time scales; evidence is heterogeneous and sometimes contradictory; and the objective function is individualized, context-dependent, and safety-critical. We argue that a few principal gaps render generic MAS inadequate for clinical decision support.

First, the classical MAS presupposes modularity, a task can be partitioned into subtasks whose interfaces are stable (e.g., routing, foraging, logistics). In clinical care, symptom clusters straddle specialties (e.g., dyspnea can span cardiology, pulmonology, allergy, rheumatology), and causal structure is uncertain ex ante. Decompositions that look tidy offline often fail at the bedside, where newly arriving information re-wires the graph of relevance. Heuristics like ``one agent per data type'' or ``one agent per ICD block'' produce brittle boundaries that do not reflect how clinicians reason across hypotheses.

Second, health data is non-stationary, and evidence can become entangled. For example, patient trajectories evolve as labs return, treatments are administered, and physiology adapts. MAS coordination policies that assume stationary payoff matrices or fixed inter-agent reliabilities drift when the data-generating process changes during a single encounter. Moreover, evidence sources (notes, imaging, medications, labs) are entangled: they are not conditionally independent given diagnosis, so naive late fusion can overweight redundant signals while under-weighting rare but decisive modalities.

Third, individual-level reliability is often under-specified. Most MAS coordination strategies optimize population-level reward or team-average success. In medicine, a system that improves the average yet fails unpredictably on particular patients is unacceptable. Group fairness constraints help at the protected-attribute level, but cannot guarantee bounded error for an arbitrary \(x_i\) in a low-density region of the joint feature space. Existing MAS rarely estimate local competence per agent for the individual case; they rarely defer or abstain when uncertainty spikes; and they seldom provide calibrated \emph{per-patient} reliability.

Finally, generic MAS rarely addresses clinical constraints, such as latency budgets for emergencies, on-device computation, audit trails for post-hoc review, or change control under adaptive learning. Safety cases demand \emph{traceable} decisions, not just good point estimates. Without structured logging, human-over-the-loop arbitration, and progressive disclosure of rationale, MAS fail institutional review, quality, and regulatory expectations.


\section{The N-of-1 AI Ecosystem}
To be suitable for care, MAS must move beyond generic coordination. They must estimate \emph{local} reliability, expose and use uncertainty, preserve dissent, allow abstention, defer to humans under critical discord, and operate within clinical MLOps and regulatory guardrails.\cite{ref17,ref18,ref19} 
We propose an ecosystem designed for \emph{individual reliability}. Rather than a monolithic predictor or a generic MAS, the system orchestrates domain-specialized agents around a shared toolkit and a coordination layer that is explicitly \emph{patient-conditional}. The aim is not merely higher average accuracy but bounded, transparent behavior for the case at hand.

\subsection{Architectural Overview}
The N-of-1 ecosystem comprises three tiers:
\begin{enumerate}
  \item \textbf{Shared Model \& Methods Repository} (foundation): a curated library of predictors, anomaly/outlier detectors, uncertainty estimators, and evidence-synthesis tools with versioning and validation metadata.
  \item \textbf{Profession-Specialized Agents} (middle): agents aligned to clinical domains (e.g., cardiology, oncology, geriatrics), each capable of selecting and adapting tools from the repository and producing \emph{probabilities, uncertainties, and rationales/provenance}.
  \item \textbf{Coordination Layer} (top): a broker that estimates each agent's \emph{local competence} for the current patient, fuses predictions with dissent-aware aggregation, enforces abstention policies, and manages human oversight.
\end{enumerate}

The Shared Model and Methods Repository functions like a hospital formulary for computation, providing various tools and methods for different tasks. It includes Predictors, which encompass population-trained baselines, subpopulation specialists, and temporal models. Atypicality Detectors such as density estimators, conformal nonconformity scores, and domain-shift indicators help assess the distance from training support. Uncertainty Estimators offer calibrated probabilities (like isotonic/Platt), ensemble variance, and conformal intervals to measure prediction uncertainty. Lastly, Evidence Synthesizers provide retrieval-augmented summaries, complete with citation provenance and conflict highlighting, to synthesize and present information effectively.

All artifacts carry metadata: data scope, validation metrics, calibration diagnostics, known failure modes, and change logs. Agents compose these tools rather than re-implement them, enabling coherent updates and governance.

The Profession-Specialized Agents act as domain-specific interpreters, tailored to their respective fields. Each agent selects the appropriate tools from the repository based on the problem at hand and the available modalities. It then contextualizes the outputs using domain-specific priors, such as treatment effects or contraindications. Finally, the agent returns a tuple  \((\hat{p}_j, u_j, R_j)\), which includes the calibrated probability, an uncertainty summary, and a rationale with provenance—such as citations, model versions, and the input features used in the process.

Agents are evaluated not only by global AUC/accuracy but by \emph{local competence curves}: performance as a function of density/shift, comorbidity strata, and modality completeness.

\subsection{Coordination Layer: Detect--Route--Defer}
Let \(x_i\) denote the current patient's feature/multimodal summary. The coordination layer computes:
\begin{align}
w_j(x_i) &= \alpha\,\underbrace{\mathrm{dens}_j(x_i)}_{\text{support}}
\;+\; \beta\,\underbrace{\mathrm{cons}_j(x_i)}_{\text{consensus}}
\;+\; \gamma\,\underbrace{\mathrm{cal}_j(x_i)^{-1}}_{\text{well-calibrated}}
\;+\; \delta\,\underbrace{\mathrm{perf}_j(x_i)}_{\text{local history}}, \label{eq:weights}
\\[4pt]
\hat{p}(x_i) &= \sigma\!\Big(a_0 + a_1\cdot \mathrm{Atyp}(x_i) + a_2\cdot \mathrm{Disagree}(x_i) + \sum_j b_j\, w_j(x_i)\,\hat{p}_j(x_i)\Big), \label{eq:stack}
\end{align}
where:
\(\mathrm{dens}_j(x_i)\) is a density score for agent \(j\)'s feature space (e.g., \(k\)-NN or conformal coverage);\\
\(\mathrm{cons}_j(x_i)\) measures alignment with other agents, weighted by domain-overlap priors;\\
\(\mathrm{cal}_j(x_i)\) summarizes recent calibration error for similar patients (lower is better);\\
\(\mathrm{perf}_j(x_i)\) captures local performance history on similar cases (recency-weighted);\\
\(\mathrm{Atyp}(x_i)\) is an atypicality score (e.g., Mahalanobis or nonconformity);\\
\(\mathrm{Disagree}(x_i)=\mathrm{sd}\{\hat{p}_j(x_i)\}\) is inter-agent dispersion;\\
\(\sigma\) is the logistic link for a simple stacker that \emph{learns} to balance agents and meta-features on validation data.

\paragraph{Routing and Abstention.}
If a high-value modality is present (e.g., \(x_3\) for a rare mechanism), a dedicated specialist can be granted veto/priority routing. When \(\mathrm{Atyp}(x_i)\) or \(\mathrm{Disagree}(x_i)\) exceed thresholds, the system can \emph{abstain} and escalate to the clinician with a concise uncertainty packet.

\paragraph{Dissent Preservation and Human Oversight.}
Rather than collapsing to a single point estimate, the layer returns:
\[
\big(\hat{p}(x_i),\; \mathrm{CI}(x_i),\; \mathcal{E}(x_i)\big),
\]
where \(\mathrm{CI}\) is a calibrated interval (e.g., conformal) and \(\mathcal{E}\) includes the top rationales, key disagreements, and provenance. A \emph{critical-discord} detector triggers clinician authority when safety-critical divergences arise.

\subsection{Coordination Pseudocode (Reference)}
\begin{verbatim}
def coordinate_agents(patient, agent_reports, policy):
    # agent_reports: list of dicts {p_hat, uncertainty, rationale, provenance}
    # 1) compute meta-features
    atyp = atypicality(patient)              # density / shift score
    disagree = dispersion([r.p_hat for r in agent_reports])
    # 2) local reliabilities
    weights = {}
    for j, r in enumerate(agent_reports):
        weights[j] = (alpha * density_j(patient)
                      + beta * consensus_j(j, agent_reports)
                      + gamma * inv_calibration_j(patient)
                      + delta * local_perf_j(patient))
    # 3) routing rules (e.g., rare-modality specialist)
    if has_high_value_modality(patient):
        primary = pick_specialist(agent_reports)
        fused = primary.p_hat
    else:
        fused = stacker_predict(patient, agent_reports, weights, atyp, disagree)
    # 4) abstain / escalate
    if should_abstain(atyp, disagree, agent_reports, policy):
        return defer_to_clinician(packet=explain(agent_reports, weights))
    return package_output(p=fused,
                          interval=conformal_interval(patient, agent_reports),
                          explanation=explain(agent_reports, weights))
\end{verbatim}

\subsection{Clinical MLOps \& Governance Hooks}
Practical deployment of the system requires several key components to ensure reliability, safety, and performance. Latency tiers are essential, with a lightweight fast path designed for acute settings and more in-depth, progressively refined analyses when time permits. On-device options enable edge inference, supporting scenarios with protected data or limited connectivity. Structured logging is used to capture inputs, agent versions, model weights \(w_j(x_i)\), and rationales, ensuring full auditability. Change control mechanisms enforce that updates to models or repositories pass through predefined safety tests and shadow runs before deployment. Finally, validation beyond averages emphasizes robust performance assessment through tail-focused metrics, selective prediction strategies (balancing risk and coverage), and per-patient calibration monitoring.

The N-of-1 ecosystem replaces generic MAS averaging with detect--route--defer intelligence that \emph{estimates and uses} local reliability, preserves dissent, abstains safely, and centers the individual patient while meeting clinical and regulatory constraints.

\section{Methods (Simulation)} 
\noindent\textbf{Rationale.}
We conducted a controlled simulation to test a claim that is difficult to isolate in observational data: population-optimized models can look strong on \emph{average} while failing precisely in low-density, atypical regions and in rare mechanisms. Real-world datasets confound this question with selection bias, unmeasured covariates, label noise, and shifting casemix, making it hard to attribute errors to model design versus data artifacts. A simulation lets us fix the ground truth data-generating process, dial the prevalence and strength of signals, and create \emph{partial observability} (signals available only to some patients) in order to stress-test decision policies.

\medskip
\noindent\textbf{Design goals.}
Our goals were to (i) construct a setting where an “average” monolithic model performs well globally yet degrades in predefined tails; (ii) embed a rare cohort with a high-value measurement that the average model cannot reliably use; (iii) evaluate whether a detect–route–defer \emph{multi-agent} design improves individual reliability without sacrificing population performance; and (iv) quantify results with tail-aware metrics (AUC/accuracy in the tail), calibration “in the small,” selective prediction (risk--coverage), and statistical tests (DeLong; paired bootstrap). Synthetic data also ensures full reproducibility, transparent ablations, and unambiguous interpretation of causal mechanisms.

\subsection{Synthetic Data Generation and Analysis}
We simulated a heterogeneous population with three \emph{majority} clusters (A--C) and one \emph{rare} cluster (D). Each individual has core covariates $(x_1,x_2)$; only cluster D additionally observes an auxiliary measurement $x_3$ (missing for A--C), which strongly drives outcomes in D.

\paragraph{Cluster geometry.}
   .\\A, B, C:\; $(x_1,x_2) \sim \mathcal{N}(\mu_c,\Sigma_{\mathrm{maj}})$ with means $\mu_A=(-1.2,1.1)$, $\mu_B=(1.4,1.0)$, $\mu_C=(0.1,-1.3)$ and
  \[
  \Sigma_{\mathrm{maj}}=\begin{pmatrix}1.0 & 0.25\\[2pt]0.25 & 1.0\end{pmatrix}.
  \] \\
  D (rare): $(x_1,x_2) \sim \mathcal{N}((2.8,-2.6),\Sigma_{\mathrm{rare}})$ with
  \[
  \Sigma_{\mathrm{rare}}=\begin{pmatrix}0.35 & -0.2\\[2pt]-0.2 & 0.5\end{pmatrix}.
  \] \\
  $x_3 \sim \mathcal{N}(0,1)$ only for cluster D; $x_3$ is \emph{missing} for A--C.

\paragraph{Outcome model.}
Let $\sigma(z)=1/(1+e^{-z})$.
\begin{align*}
\text{Majority (A--C):}\quad & \Pr(Y{=}1\mid x_1,x_2)=\sigma\!\left(0.8+1.2x_1-0.9x_2+0.35\,x_1x_2\right).\\
\text{Rare (D):}\quad & \Pr(Y{=}1\mid x_3)=\sigma(\beta_3 x_3),\;\; \beta_3=3.5.
\end{align*}
To mimic label noise outside D we apply a small symmetric flip probability to A--C:
$p \leftarrow p(1-0.03) + (1-p)\,0.03$; for D we set the flip rate to 0. Outcomes are $Y\sim \mathrm{Bernoulli}(p)$.

\paragraph{Sample sizes and split.}
We simulated $N=12{,}400$ individuals (A=6000, B=4000, C=2000, D=400) and performed a \emph{stratified} split by cluster into train/validation/test of $60\%/20\%/20\%$ (per cluster).

\subsection*{Tail definition and local density}
Atypicality is quantified in two ways:
\begin{enumerate}
  \item \textbf{Mahalanobis tail (primary).} Compute $m(x)=\mathrm{MD}(x;\hat\mu_{\mathrm{train}},\hat\Sigma_{\mathrm{train}})$ in $(x_1,x_2)$ and define the \textbf{tail} as the top $12\%$ of test points by $m(x)$.
  \item \textbf{Local density (secondary).} For plots/meta-features use $\rho(x)=1/(d_k(x){+}\epsilon)$, the inverse distance to the $k$-th nearest training neighbor with $k=25$ (smaller $\rho$ $\Rightarrow$ more atypical).
\end{enumerate}

\subsection{Models}
\paragraph{Monolithic baseline (``average'' model).}
Because $x_3$ is largely missing, the monolith uses only $(x_1,x_2)$ and ensembles:
(i) quadratic logistic regression $\text{logit}(Y) \leftarrow \text{poly}(x_1,2)+\text{poly}(x_2,2)+x_1x_2$ and
(ii) a random forest (600 trees) on $(x_1,x_2)$. The final score is the simple average of the two probabilities.

\paragraph{Multi-agent system.}
\begin{enumerate}
  \item \textbf{Regional agents.} Cluster the training $(x_1,x_2)$ space with $K{=}3$ $k$-means. For each region, select \emph{GLM vs RF} by 3-fold CV AUC and refit the winner on that region (only $(x_1,x_2)$).
  \item \textbf{Rare-case specialist.} On the subset with observed $x_3$ (cluster D) fit $Y\sim \text{logit}^{-1}(\alpha+\gamma x_3)$. At inference this specialist is \emph{only} queried when $x_3$ is available.
  \item \textbf{Meta-learner (stacking).} On validation, compute agent probabilities $\{p_j(x)\}$ and two meta-features: local density $\rho(x)$ and agent disagreement $\mathrm{sd}\{p_j(x)\}$. Fit a logistic stacker
  \[
  \Pr(Y{=}1\mid x)=\text{logit}^{-1}\!\left(a_0+a_1\rho(x)+a_2\,\mathrm{sd}\{p_j(x)\}+\sum_j b_j p_j(x)\right),
  \]
  with \textbf{tail up-weighting} ($\times 4$) for low-density points ($<$ 8th percentile).
  \item \textbf{Routing.} If $x_3$ is present, \emph{defer} to the $x_3$-specialist; otherwise use the stacked probability. (Ablations also evaluate proximity/equal agent weighting without stacking or specialist.)
\end{enumerate}

\subsection{Evaluation}
Primary metrics are ROC AUC and accuracy at threshold 0.5, reported overall, on the 12\% Mahalanobis tail, and per cluster (A--D). AUCs are compared with the two-sided DeLong test for correlated ROC curves; 95\% CIs are reported from the ROC procedure. Tail calibration is summarized by Expected Calibration Error (ECE) on quantile bins. Selective prediction uses risk--coverage curves under a simple confidence policy (if $x_3$ present $\Rightarrow$ high confidence; else by $|p-0.5|$). Robustness is assessed by a paired bootstrap on the test set ($B{=}1000$) that recomputes the tail within each resample and reports the 2.5/50/97.5 percentiles and two-sided $p$ for deltas.

\subsection*{Reproducibility}
All simulations were implemented in \textsf{R} with fixed seeds (\texttt{set.seed(2025)} for data generation; fixed seeds for splits and model selection). Train/validation/test split was stratified by cluster at 60\%/20\%/20\%. Tail share was pre-specified at 12\% of the test set by Mahalanobis distance in $(x_1,x_2)$. Local density used $k=25$ nearest neighbors to the training set. Number of agents $K=3$ (k-means); RF trees $=600$; tail up-weighting factor in the stacker $=4$; confidence rule for risk--coverage: high confidence if $x_3$ present, else by $|p-0.5|$. AUC comparisons used the two-sided DeLong test; 95\% CIs followed pROC defaults. Bootstrap used $B=1000$ paired resamples and recomputed the tail subset within each resample.

\section{Results}

\begin{figure}[t]
  \centering
  \includegraphics[width=\textwidth]{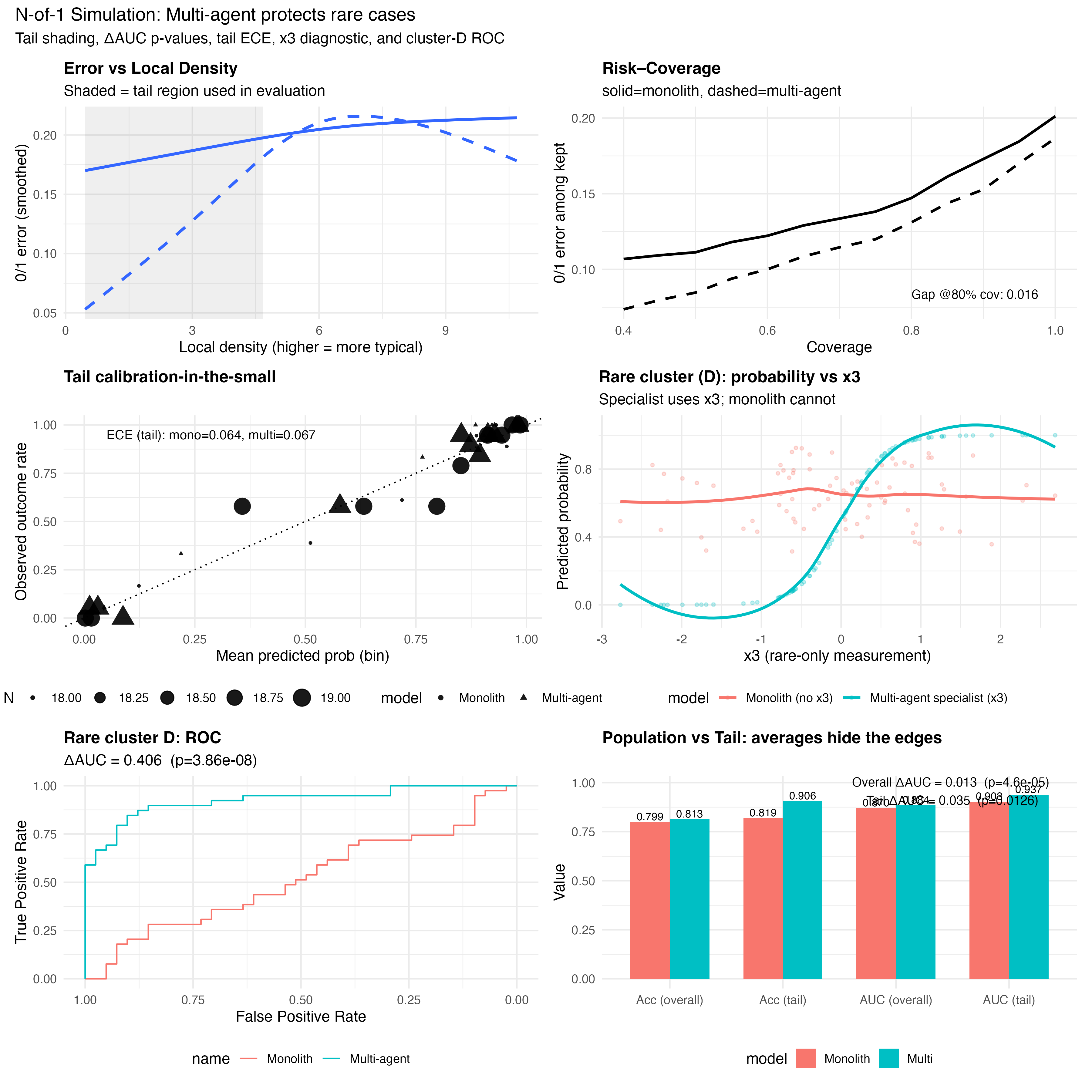}
  \caption{Summary panel: error vs.\ density (tail shaded), risk--coverage (solid = monolith; dashed = multi-agent), tail calibration (ECE), $x_3$ diagnostic for cluster D, ROC for cluster D ($\Delta$AUC = 0.406; $p\approx 3.9\times 10^{-8}$), and population vs.\ tail bars (overall $\Delta$AUC = 0.013; tail $\Delta$AUC = 0.035).}
  \label{fig:panel}
\end{figure}

Simulation results on the full test set showed that, the multi-agent system outperformed the monolithic baseline, where AUC increased from \textbf{0.870} to \textbf{0.884} ($\Delta$AUC=\textbf{0.013}, $p=4.6\times10^{-5}$) and accuracy from \textbf{0.799} to \textbf{0.813} (Table~\ref{tab:summary}; Figure~\ref{fig:panel}).

\paragraph{Special/rare cases (tail).} In the pre-specified tail (top 12\% by Mahalanobis distance) the gap widened: AUC \textbf{0.903} $\rightarrow$ \textbf{0.937} ($\Delta=\mathbf{0.035}$, $p=\mathbf{0.0126}$) and accuracy \textbf{0.819} $\rightarrow$ \textbf{0.906}. This shows that population averages understate material errors in atypical patients (Table~\ref{tab:summary}; Figure~\ref{fig:panel}).

\paragraph{Rare cluster D (N-of-1 mechanism).} For the rare cohort D, where $x_3$ is observed and the system defers to a specialist, performance jumped from AUC \textbf{0.518} to \textbf{0.924} ($\Delta=\mathbf{0.406}$, $p=\mathbf{3.86\times10^{-8}}$) and from accuracy \textbf{0.462} to \textbf{0.875} (Table~\ref{tab:summary}). The $x_3$ diagnostic and D-specific ROC in the panel confirm the mechanism.

\paragraph{Selective prediction \& calibration.} Under the simple confidence policy, the multi-agent risk--coverage curve dominates; at 80\% coverage the error among kept points is lower by \textbf{0.016} (Figure~\ref{fig:panel}, top-right). Tail calibration is comparable (ECE: monolith \textbf{0.064} vs multi-agent \textbf{0.067}) while discrimination improves (Figure~\ref{fig:panel}, bottom-left).

\paragraph{Ablations \& robustness.} Removing the \emph{rare-case specialist} produces the largest loss on tail metrics; removing stacking yields modest changes relative to the full model; agents-only underperforms the full system on tail AUC (Table~\ref{tab:ablation}). Paired bootstrap CIs for deltas remain positive and statistically significant (Table~\ref{tab:bootstrap}).

\begin{table}[t]
\centering
\small
\setlength{\tabcolsep}{3.5pt}  
\caption{Summary metrics (AUC and Accuracy) by setting.}
\label{tab:summary}
\begin{tabular}{@{} L{32mm} l c c L{24mm} @{}}
\toprule
Setting & Metric & Monolith & Multi-agent & $\Delta$ (p) \\
\midrule
Overall           & AUC & 0.870 & 0.884 & 0.013 ($4.6\times10^{-5}$) \\
Overall           & Acc & 0.799 & 0.813 & 0.015 (—) \\[3pt]
Tail (top 12\%)   & AUC & 0.903 & 0.937 & 0.035 (0.0126) \\
Tail (top 12\%)   & Acc & 0.819 & 0.906 & 0.087 (—) \\[3pt]
Rare cluster D    & AUC & 0.518 & 0.924 & 0.406 ($3.86\times10^{-8}$) \\
Rare cluster D    & Acc & 0.462 & 0.875 & 0.412 (—) \\
\bottomrule
\end{tabular}
\end{table}

\begin{table}[t]
\centering
\caption{Per-cluster performance on the test set.}
\label{tab:percluster}
\begin{tabular}{lccccc}
\toprule
Cluster & $N$ & AUC (Mono) & AUC (Multi) & Acc (Mono) & Acc (Multi) \\
\midrule
A & 1200 & 0.815 & 0.820 & 0.800 & 0.796 \\
B &  800 & 0.746 & 0.763 & 0.816 & 0.821 \\
C &  400 & 0.646 & 0.620 & 0.828 & 0.838 \\
D &   80 & 0.518 & 0.924 & 0.463 & 0.875 \\
\bottomrule
\end{tabular}
\end{table}

\begin{table}[t]
\centering
\caption{Ablation study (same TEST split).}
\label{tab:ablation}
\begin{tabular}{lcccc}
\toprule
Model & AUC (overall) & ACC (overall) & AUC (tail) & ACC (tail) \\
\midrule
Monolith       & 0.870 & 0.799 & 0.903 & 0.819 \\
Multi\_full    & 0.884 & 0.813 & 0.937 & 0.906 \\
No\_specialist & 0.876 & 0.801 & 0.903 & 0.822 \\
No\_stacking   & 0.885 & 0.819 & 0.947 & 0.906 \\
Agents\_only   & 0.870 & 0.804 & 0.901 & 0.826 \\
\bottomrule
\end{tabular}
\end{table}

\begin{table}[t]
\centering
\caption{Paired bootstrap on TEST ($B=1000$): deltas and CIs.}
\label{tab:bootstrap}
\begin{tabular}{lcccc}
\toprule
Metric & 2.5\% & Median & 97.5\% & $p$ (paired bootstrap) \\
\midrule
$\Delta$AUC overall & 0.008 & 0.013 & 0.020 & 0 \\
$\Delta$ACC overall & 0.005 & 0.015 & 0.024 & 0.008 \\
$\Delta$AUC tail    & 0.008 & 0.033 & 0.062 & 0.014 \\
$\Delta$ACC tail    & 0.050 & 0.084 & 0.124 & 0 \\
\bottomrule
\end{tabular}
\end{table}

\section{Discussion}

The idea of an N-of-1 AI ecosystem cannot rest on metaphor or architectural elegance alone. Medicine, unlike engineering, is a field where failures are counted in lives, not in logs. To be worthy of adoption, the system must demonstrate its value through rigorous validation, show that it can be deployed within the constraints of clinical practice, and carry safeguards strong enough to catch its own missteps.\cite{ref20,ref21}

The coordination layer may be the conductor, but medicine is not a self-playing symphony. The final audience and the final arbiter is the clinician. This ecosystem was never designed to replace their judgment; it was built to refine it.

Consider the ordinary rhythm of a clinic visit: the physician reviews vitals, scans, labs, and patient history. Each fragment of data is incomplete, shaped by chance and circumstance. What the N-of-1 ecosystem offers is not more fragments, but coherence. A cardiology agent may highlight a gradual decline in ejection fraction, while the geriatrics agent flags polypharmacy invisible to a population-trained model. The coordination layer harmonizes these voices and then packages the result into something legible, a risk estimate with uncertainty bounds, a warning that “this patient looks unlike most of the data,” and curated evidence snippets that anchor recommendations in published knowledge.

For the clinician, this is not an impenetrable black box but a kind of augmented stethoscope, tuned to patterns they could not hear unaided. Crucially, the design avoids overwhelming the user. Rather than flooding the screen with agent chatter, it distills their collective reasoning into a packet: a summary risk profile, context for why the prediction deviates, and options for action. 

Importantly, symbiosis is safeguarded by an arbitration logic that recognizes the limits of computation. When critical discord emerges between agents, or between agents and clinicians, the protocol defaults to clinician authority. This ensures that coordination does not collapse into automation, but remains a layered conversation where machine contributions are modulated, not absolute.

This symbiosis respects the clinician’s primacy. The AI does not dictate; it suggests. It does not silence disagreement among agents; it surfaces it. In moments of clarity, it accelerates decisions. In moments of uncertainty, it warns that the path ahead is foggy. The machine’s role is not to be the physician, but to ensure the physician never stands alone with partial information.

If population-trained models embody the average, this ecosystem restores the singular: the individual patient, in all their specificity, complexity, and irreducible difference. In doing so, it preserves what medicine has always sought, not the care of populations in abstract, but the care of individuals. While determining the appropriate balance between aggregation and disaggregation remains a central issue in algorithmic and statistical inference, our architecture offers a framework for navigating this trade-off, ensuring that the most precise outcomes are achieved when necessary.\cite{ref22,ref23,ref24}

The first question is deceptively simple: how will we know if the system works? The usual measures, area under the curve, and overall accuracy, were built for the logic of populations. They can hide the very failures that matter most here: those that occur on the edges, where the patient is not like the others.

A sharper lens is needed. Outlier-aware metrics such as calibration slope in distant subgroups, or reductions in false negatives among patients lying farthest from the population mean (for example, those identified by Mahalanobis distance), tell us whether the system bends toward the atypical. Validation, too, must follow a sequence. First, in silico tests on vast multi-center datasets, where the rare becomes visible. Next, trials with synthetic “digital twins” that exaggerate edge cases to probe resilience. Finally, carefully bound prospective pilots in clinics, where the demands of time, trust, and context bring theory into contact with reality. Each stage serves as a sieve, allowing only models that truly improve care for the hard cases to pass through.

The architecture is ambitious: multiple agents, each with specialized knowledge, orchestrating in real time. The question clinicians will ask is not whether it is clever, but whether it can keep up. In a controlled lab, the machinery hums at leisure. In an emergency room, minutes can mean myocardium.

Practicality must be baked in. Agents cannot afford to recompute everything from scratch. They must cache models, optimize queries, and prune redundancies. For life-or-death contexts, stroke or cardiac arrest, the system must be able to push essential functions to the “edge,” running on local hardware with latency measured in milliseconds. Scalability here is not a matter of elegance but of survival.

Every architecture has seams where it can split. Agents may drift into silos, outlier detectors may confuse noise for signal, and language models may fabricate citations. These are not hypothetical flaws; they are the predictable costs of complexity. To ignore them would be negligent.

Safeguards, therefore, must be structural. Confidence thresholds should govern which outputs reach a clinician’s eyes. Cross-agent consensus checks ensure that no single model is granted unexamined authority. Most of all, the human remains the final arbiter for high-stakes decisions: admission, surgery, therapy. The ecosystem is scaffolding, not replacement. Its worth lies in making uncertainty explicit, in giving clinicians the means to see disagreement rather than smoothing it away.

An N-of-1 ecosystem cannot stand on architecture alone. For it to move from theory to clinic, it must earn legitimacy through regulation, trust through transparency, and durability through ethics. The pathway is less a straight corridor than a gauntlet of checkpoints, scientific, institutional, and societal.

Traditional regulatory models were built for fixed entities: a pill with a stable dose, a device with a predictable range. But the N-of-1 ecosystem is neither fixed nor static. It shifts with each patient, reconfiguring itself around the contours of an individual life. Such adaptability strains against regulatory systems that prize uniformity. A new framework must be layered: retrospective evaluations to establish baseline safety; digital twin simulations to probe performance under thousands of synthetic variations; and carefully bounded prospective pilots in live clinics. The final stage, post-market surveillance, becomes not an addendum but the linchpin, continuously testing the system where it is weakest: on the margins, with the outliers.

Beyond population metrics, validation must become individualized and multi-modal. Although the challenge goes beyond mistaking correlation for causation,.\cite{ref25} synthetic counterfactuals (“digital twins”) allow us to test whether the system can navigate edge cases; per-patient calibration intervals monitor predictive stability over time; and shadow mode deployments offer a parallel stream of recommendations without disrupting care. For chronic patients, N-of-1 trials can alternate AI-assisted and standard-care phases, yielding individualized evidence. These methods ensure that the system remains fair by addressing disparities across diverse patient groups, ensuring that the model's predictions are accurate and equitable for each individual. By focusing on personalized validation techniques, we can identify and mitigate biases, delivering care that is not only effective but also just for all patients.\cite{ref26,ref27,ref28} 

We also propose a “surprise index”, a measure of when outcomes diverge sharply from predictions, as a safeguard against unrecognized blind spots. Together, these practices reframe validation as an ongoing relationship with the patient rather than a one-time population benchmark.

\section{Conclusion}
Medicine has always wrestled with averages. The “standard dose,” the “reference range,” the “typical presentation”, all are useful fictions that smooth the jaggedness of human variation. Artificial intelligence, for all its sophistication, has inherited this same weakness. It excels in the middle of the curve, yet falters where medicine matters most: on the margins, with the patient who does not conform.

We have called this the average patient fallacy, and it is more than a technical oversight. It is a fracture in the moral contract of medicine. To serve only the average is to abandon the very people who need precision most, the rare, the underrepresented. 

The framework we propose, a distributed ecosystem of specialized agents, orchestrated toward N-of-1 decision support, offers one way forward. It is not a promise of flawless prediction, but of a system designed to notice, to contest, and to adapt. It is an architecture that acknowledges fallibility, yet bends toward fidelity with the patient in front of us. This is not a blueprint for perfection but for progress. It asks us to imagine AI not as an oracle but as a colleague: distributed, disputatious, and deeply human in its fallibility. The shift is subtle but profound, from a system that optimizes for the population to one that learns to care for the individual.

The measure of success will not be in the elegance of its algorithms, but in the quiet moments where it alters a decision, spares a delay, or changes a life. In shifting from the population to the person, medical AI may finally begin to honor the very essence of clinical care.

Individualization deepens the stakes for privacy. To tailor decisions around the singular patient is also to risk exposing their singularity. Technical safeguards are necessary, federated learning to keep raw data in place, encrypted updates to shield patient identifiers, and local computation for sensitive tasks. Yet trust will not be secured by code alone. Clinicians and patients need a window into how the system learns and acts, a line of sight into governance and accountability. Without that transparency, the ecosystem becomes indistinguishable from the black boxes it seeks to replace.

The average patient fallacy is not just a mathematical distortion; it is a quiet betrayal. By optimizing for the many, we fail those who do not conform, the rare and  complicated. An N-of-1 system inverts that logic. It treats deviation not as an error to be suppressed but as a signal to be studied. Its ethical commitment lies in acknowledging that medicine’s deepest duty is not to the aggregate curve but to the individual at its edge. In such a system, fairness is not an accessory; it is the architecture.

\section*{Acknowledgments}
This study has been supported by grants from the National Institutes of Health: The National Institute on Aging R01AG074372 and The National Institute of Allergy and Infectious Diseases R01AI165535.

\section*{Code \& Data Availability}
All code for generating the synthetic data, training models, and reproducing figures/tables is available in the project repository (synthetic data only). The public project repository can be found here: \url{https://github.com/MGH-CLAI/NOf1}.

\section*{Ethics}
This study used synthetic data only and did not involve human subjects or protected health information.

\end{document}